\documentclass[letterpaper]{article} 
\usepackage{aaai25}  
\usepackage{times}  
\usepackage{helvet}  
\usepackage{courier}  
\usepackage[hyphens]{url}  
\usepackage{graphicx} 
\usepackage{natbib}  
\usepackage{caption} 
\usepackage{algorithm}
\usepackage{algorithmic}

\usepackage{newfloat}
\usepackage{listings}
\DeclareCaptionStyle{ruled}{labelfont=normalfont,labelsep=colon,strut=off} 
\floatstyle{ruled}
\newfloat{listing}{tb}{lst}{}
\floatname{listing}{Listing}
\usepackage{graphicx}
\usepackage{booktabs}
\usepackage{adjustbox}
\usepackage{multirow}
\usepackage{amsmath}
\usepackage{amssymb}

\title{Improving Concept Alignment in Vision-Language Concept Bottleneck Models}
\author{
    Nithish Muthuchamy Selvaraj, 
    Xiaobao Guo, 
    Adams Wai-Kin Kong, 
    Alex Kot
}
\affiliations{
    Nanyang Technological University, Singapore\\
}

\begin{document}

\maketitle


\begin{abstract}
Concept Bottleneck Models (CBM) map images to human-interpretable concepts before making class predictions. Recent approaches automate CBM construction by prompting Large Language Models (LLMs) to generate text concepts and employing Vision Language Models (VLMs) to score these concepts for CBM training. However, it is desired to build CBMs with concepts defined by human experts rather than LLM-generated ones to make them more trustworthy. 
In this work, we closely examine the faithfulness of VLM concept scores for such expert-defined concepts in domains like fine-grained bird species and animal classification. Our investigations reveal that VLMs like CLIP often struggle to correctly associate a concept with the corresponding visual input, despite achieving a high classification performance. This misalignment renders the resulting models difficult to interpret and less reliable.
To address this issue, we propose a novel Contrastive Semi-Supervised (CSS) learning method that leverages a few labeled concept samples to activate truthful visual concepts and improve concept alignment in the CLIP model. Extensive experiments on three benchmark datasets demonstrate that our method significantly enhances both concept ($+29.95\%$) and classification ($+3.84\%$) accuracies yet requires only a fraction of human-annotated concept labels.
To further improve the classification performance, we introduce a class-level intervention procedure for fine-grained classification problems that identifies the confounding classes and intervenes in their concept space to reduce errors. Source codes are available at \url{https://github.com/NMS05/Improving-Concept-Alignment-in-Vision-Language-Concept-Bottleneck-Models}.

\end{abstract}


\vspace{-0.5em}
\section{Introduction}
\label{sec:intro}

Concept Bottleneck Models (CBM)~\cite{koh2020concept} are a popular choice to build interpretable models using human-understandable concepts. These models first map input images to a low-dimensional, expert-defined concept space. For instance, the bird species `Vermilion Flycatcher' can be described using concepts like `red-body' and `black-wing'. A linear classifier then makes predictions based on these concepts. Building a CBM requires a well-defined, non-ambiguous set of concepts and the corresponding concept labels for each sample, which is labor-intensive to obtain. Recent approaches~\cite{oikarinenlabel,yang2023language} circumvent this issue by automating CBM construction using Large Language Models (LLMs) and Vision Language Models (VLMs). These methods generate an initial concept set by prompting an LLM with few-shot examples, then extracting a subset of those concepts that are conducive to classification. When an image and this filtered concept set are input to a contrastively pre-trained VLM, its image-text alignment scores serve as ``concept scores" or ``concept labels" for that sample. These scores can be used to train another CBM classifier or directly predict the classes, effectively turning a VLM into a CBM (referred to as VL-CBM hereon).

While LLMs possess rich world knowledge~\cite{petroni2019language,jiang2020can} and can generate extensive concept sets across various domains, expert-defined concepts are sometimes preferred for several reasons.
\begin{itemize}
    \item \textbf{Concise Concept Sets:} A small, well-defined set of expert concepts allows for easier error analysis and intervention, which is crucial for interpretable models.
    \item \textbf{Trusted Expertise:} In medical domains like Hematology~\cite{tsutsui2023wbcatt}, expert concepts are preferred over LLM-generated ones, as the former comes from the more trusted pathologists.
    \item \textbf{Visual Grounding:} For faithful visual classification, all concepts must be visually grounded in the image, which is ensured with expert concepts.
\end{itemize}

In such cases with limited yet preferable expert concepts, it is essential to ask: \textit{Do VLM concept scores faithfully represent the visual input, and are they effective for building VL-CBMs with the desired classification performance?} To answer this, we closely examine CLIP~\cite{radford2021learning} concept scores for expert concepts on classification datasets such as CUB~\cite{wah2011caltech}, RIVAL~\cite{moayeri2022comprehensive}, and AwA2~\cite{xian2018zero}. Ideally, the CLIP model should accurately associate visual concepts (appearance, color, shape, etc.) with the corresponding visual input. However, our investigations reveal two key issues:
\begin{itemize}
    \item \textbf{Poor Concept Alignment:} The CLIP model exhibits relatively lower concept accuracy despite the higher classification performance of VL-CBMs. This suggests that CLIP's concept scores do not faithfully represent the visual input, thus making the resulting model less reliable.
    \item \textbf{Challenges with Fine-Grained Concepts:} For challenging classification tasks like CUB, the CLIP model tends to be biased towards the primary color of the bird, incorrectly attributing this dominant color to other body parts. It struggles to correctly attribute the fine-grain concepts to the visual input.
\end{itemize}

Hence, \textit{to improve the faithfulness and reliability of VL-CBMs, there is a need to improve concept alignment \textit{i.e.,} activate the desired/truthful concepts for a given image, which requires concept supervision.} However, obtaining supervisory concept labels for all training samples is cumbersome. 
To address this, we propose a novel Contrastive Semi-Supervised (CSS) learning approach that improves concept alignment in VL-CBM with fewer labeled concept examples. Our approach encourages consistent concept scores for intra-class samples while contrasting them with other classes. It then aligns these scores with ground truth using a few labeled examples per class (semi-supervision). Extensive experiments on three benchmark datasets CUB, RIVAL, and AwA2 show that our CSS method substantially increases the concept accuracy (+39.1\% for CUB, +18.63\% for RIVAL, +32.11\% for AwA2) and enhances the overall classification accuracy (+5.61\% for CUB, +2.84\% for RIVAL, +3.06\% for AwA2) with minimal human-annotated concept labels. We also validate the effectiveness our CSS approach in a medical domain using the blood cell attribute dataset WBCAtt~\cite{tsutsui2023wbcatt}.

\textbf{Intervention} is the key advantage of a CBM over other interpretable models, where incorrect predictions of the classifier can be rectified by examining and modifying the concepts. In this work, we propose a class-level intervention procedure for fine-grain classification problems, where we first compute an error matrix to identify the ``confounding classes'' (visually similar but semantically different) for which the model makes the most error. We then intervene the error images of these classes to improve overall classification accuracy. The primary contributions of this paper are summarized below,
\begin{itemize}
    \item \textbf{Evaluating faithfulness of VL-CBMs.} We quantify and evaluate the faithfulness of VLM concept scores, revealing that CLIP models exhibit poor concept alignment and struggle with fine-grain concept association.
    \item \textbf{Improving Concept Alignment:} We introduce a novel Contrastive Semi-Supervised (CSS) approach that leverages a fraction of human-annotated and expert-defined concept labels, which are inherently more reliable, to improve both concept and classification accuracies.
    \item \textbf{Class-level Intervention:}  We propose an intervention procedure for fine-grain classification tasks that reduces errors in confounding classes, further enhancing classification performance.
\end{itemize}


\section{Related Work}


Early works on Explainable AI focused on unimodal (image) classifiers and they can be broadly classified into two categories.
\textbf{Prototype-based} methods~\cite{chen2019looks,nauta2021neural,Donnelly_2022_CVPR, wang2023learning} learn interpretable visual feature prototypes for every class from the training samples. During inference, the distance (similarity score) between the test sample feature and the learned prototypes determines the classifier's decision. 
\textbf{Concept-based} methods~\cite{kim2018interpretability,koh2020concept,zarlenga2022concept} map the input to a high-level concept space defined by human experts, where these concepts are readily interpretable. The classifier decision is then based on the predicted concept scores. The main advantage of concept-based methods is that they allow test-time intervention~\cite{mitchell2021fast,abid2021meaningfully,bontempelli2023conceptlevel} and they are easier to debug compared to other methods. 

\vspace{0.2em}
\noindent\textbf{Vision-Language interpretable models (VL-CBMs).} A limitation of concept-based models is that they require concept labels for every training sample~\cite{koh2020concept} or a lot of training images representing each of those concepts~\cite{kim2018interpretability}. 
To mitigate this, recent approaches~\cite{oikarinenlabel,yang2023language,yan2023robust} leverage contrastively pre-trained Vision-Language Models, like CLIP, to automatically generate concept labels from their image-text alignment scores. 
These models have natural language concepts~\cite{menon2022visual,yan2023learning} which make them easily interpretable.

\vspace{0.2em}
\noindent\textbf{Enhancing faithfulness.}
Though pioneering works on VL-CBMs have proposed methods to extract salient concepts for higher classification accuracy, recent studies have identified several drawbacks, such as concept prompts being biased towards object names~\cite{zang2024pre}, concepts being non-factual or irrelevant~\cite{fang2024cross}, and insufficient concept coverage~\cite{shang2024incremental}.
These recent works improve VL-CBM reliability by discovering concepts that better represent visual attributes~\cite{esfandiarpoor2024if,chiquier2024evolving} and have stronger CLIP alignment scores.
In comparison, we use expert concepts, which are inherently more trustworthy but limited in size. Importantly, our CSS method can be applied to these works provided a few labels can be annotated for those concepts.
Some works also enhance VL-CBMs' reliability by making the concepts resilient to perturbations~\cite{lai2023faithful} and using sparse classifiers~\cite{panousis2023sparse,semenov2024sparse}.


\section{Methodology}

\subsection{Vision-Language Concept Bottleneck Models}
We first formally define VL-CBMs. Consider a dataset $D=\{(x_1,y_1),(x_2,y_2),...,(x_N,y_N)\}$, where $x_i \in \mathbb{R}^{H \times W \times 3}$ denotes the images, $y_i$ denotes the labels for $k$ classes, and $N$ is the total number of training samples. While a black-box classifier $f: x \rightarrow \mathbb{R}^k$ directly maps an image $x$ to its label $y$, a CBM consists of a backbone $g: x \rightarrow \mathbb{R}^c$ which first maps the image $x$ to a concept space consisting of $c$ pre-defined concepts and a linear classifier $h: \mathbb{R}^c \rightarrow \mathbb{R}^k$ that maps the predicted concepts to its corresponding label $y$. These concepts represent the salient visual features crucial for solving the classification task and are human interpretable.

However, training the backbone $g$ requires $N \times c$ concept labels which is tedious to obtain. VL-CBMs overcome this issue by leveraging the image-text alignment scores of contrastively pre-trained VLMs (like CLIP) as concept scores. Concretely, let $T=\{t_1, t_2,...,t_c\}$ be the set of LLM generated or expert-defined text concepts. Now the $c$ concept scores for an image $x_i$, which is denoted as $C_i$, is obtained by $C_i = E_I(x_i) \cdot E_T(T)^{\intercal}$, where $\cdot$ denotes the dot product, $E_I$ and $E_T$  denotes the image and text encoder of the CLIP model that maps the images and text concepts into a shared feature space. Now, with the availability of image $x_i$, concept score $C_i$, and class label $y_i$ for $N$ samples, the linear classifier $h$ can be trained to make class predictions.


\begin{table}[t]
    \centering
    \vspace{-0.5em}
    \begin{adjustbox}{width=0.4\textwidth}
    \begin{tabular}{cccc}
        \toprule
         & \textbf{~~~CUB} & \textbf{~~~RIVAL} & \textbf{~~~AwA2} \\
        \midrule
        Classification Acc (\%) & ~~75.84 & ~~95.63 & ~~90.14 \\
        Concept Acc (\%) & ~~~24.43 & ~~~58.85 & ~~~49.02 \\
        \midrule
        No.of Concepts & ~~~312 & ~~~18 & ~~~85 \\
        Concept Difficulty & ~~~Hard & ~~~Easy & ~~~Medium \\
        \bottomrule
    \end{tabular}
    \end{adjustbox}
    \vspace{-0.5em}
    \caption{Evaluating faithfulness of CLIP concept scores and VL-CBM classification performance.}
    \label{tab: poor concept alignment}
    \vspace{-1.5em}
\end{table}


\subsection{Evaluating Faithfulness of VL-CBMs}
\begin{figure*}[t]
  \centering
  \includegraphics[width=0.75\linewidth]{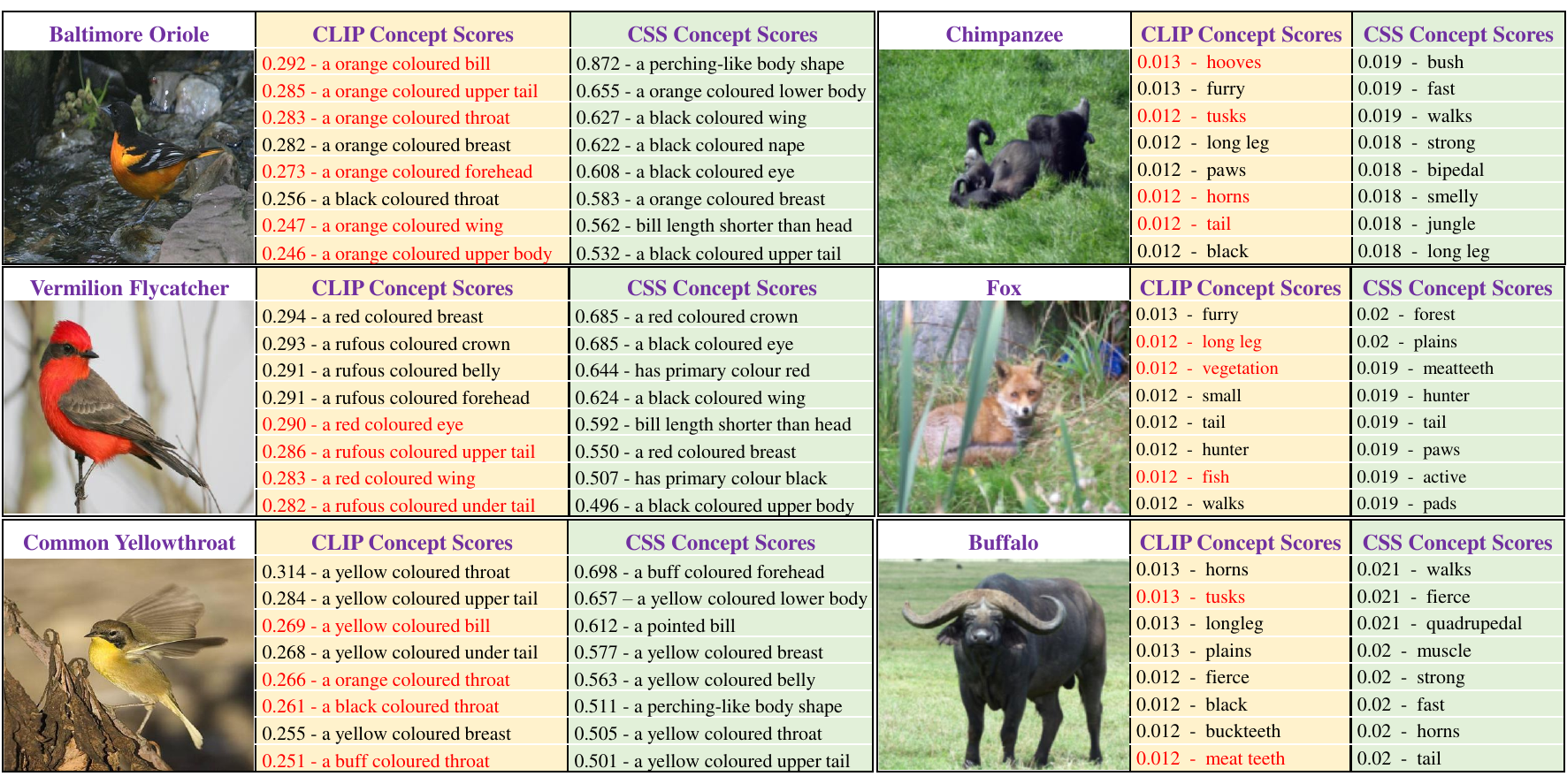}
  \vspace{-0.5em}
  \caption{Visualization of $top-8$ concept scores for the CUB (birds) and AwA2 (animals) datasets. Incorrectly activated concepts are highlighted in red.}
  \label{fig:concept visualizations}
  \vspace{-1.0em}
\end{figure*}

The trustworthiness and reliability of a VL-CBM depends on the faithfulness of its concept scores, which serve two purposes: providing rich representations for a linear classifier and enabling human interpretation of the classifier's predictions. Prior works on VL-CBMs have focused primarily on classification accuracy, which measures \textit{how well samples are separable in the concept space}, but \textit{doesn't measure the faithfulness of concept scores in representing visual truth}. Therefore, we evaluate both the faithfulness of VLM concept scores by comparing them to ground-truth concept labels~\cite{koh2020concept} and their classification performance with a linear classifier. We use three image classification datasets - Caltech-UCSD Birds (CUB)~\cite{wah2011caltech}, Rich Visual Attributes with Localization (RIVAL)~\cite{moayeri2022comprehensive}, and Animal with Attributes (AwA2)~\cite{xian2018zero} because they provide expert-defined concepts and the corresponding ground-truth concept labels needed for our evaluation. The results are presented in Table~\ref{tab: poor concept alignment} and a visualization of these concepts is shown in Figure~\ref{fig:concept visualizations}. Our investigations reveal two key issues with CLIP concept scores.

\vspace{0.2em}
\noindent\textbf{Poor Concept Alignment:} From the table, the CLIP model has relatively lower concept accuracy across all three datasets, despite the VL-CBM achieving high classification accuracy. This indicates that CLIP's concept scores do not faithfully represent the visual input and a higher classification performance does not guarantee a faithful concept understanding. Consequently, the model is less reliable, and it becomes difficult to intervene such a model during test-time.

\vspace{0.2em}
\noindent\textbf{Challenges with Fine-Grained Concepts:} A closer inspection of these concepts in Figure~\ref{fig:concept visualizations} (CLIP Concept Scores) reveals that, for challenging tasks like CUB, the CLIP model tends to be biased towards the primary color of the bird and incorrectly attributes this dominant color to other body parts. It struggles to correctly associate the compositional and fine-grain concepts with the visual input.

\vspace{0.2em}
\noindent\textbf{Why does CLIP exhibit poor concept alignment?} 
%
Although the CLIP model excels at associating images with text containing object names and attributes, achieving strong zero-shot performance on open-ended classification tasks, we find that the CLIP model struggles with fine-grained visual understanding when only concept descriptions are presented, making errors in associating objects with their attributes. The CLIP model shows unsatisfactory concept accuracy when dealing with such fine-grained concepts defined by experts. Similar findings are also discussed in some previous works~\cite{zang2024pre,pham2024peeb,yuksekgonul2022and,lewis2022does,yun2022vision}. 
As poor concept alignment can reduce the reliability and interpretability of VL-CBMs, it is critical to improve the faithfulness of VL-CBMs by activating the truthful concepts for the given image.

\begin{figure*}[t]
  \centering
  \includegraphics[width=0.65\linewidth]{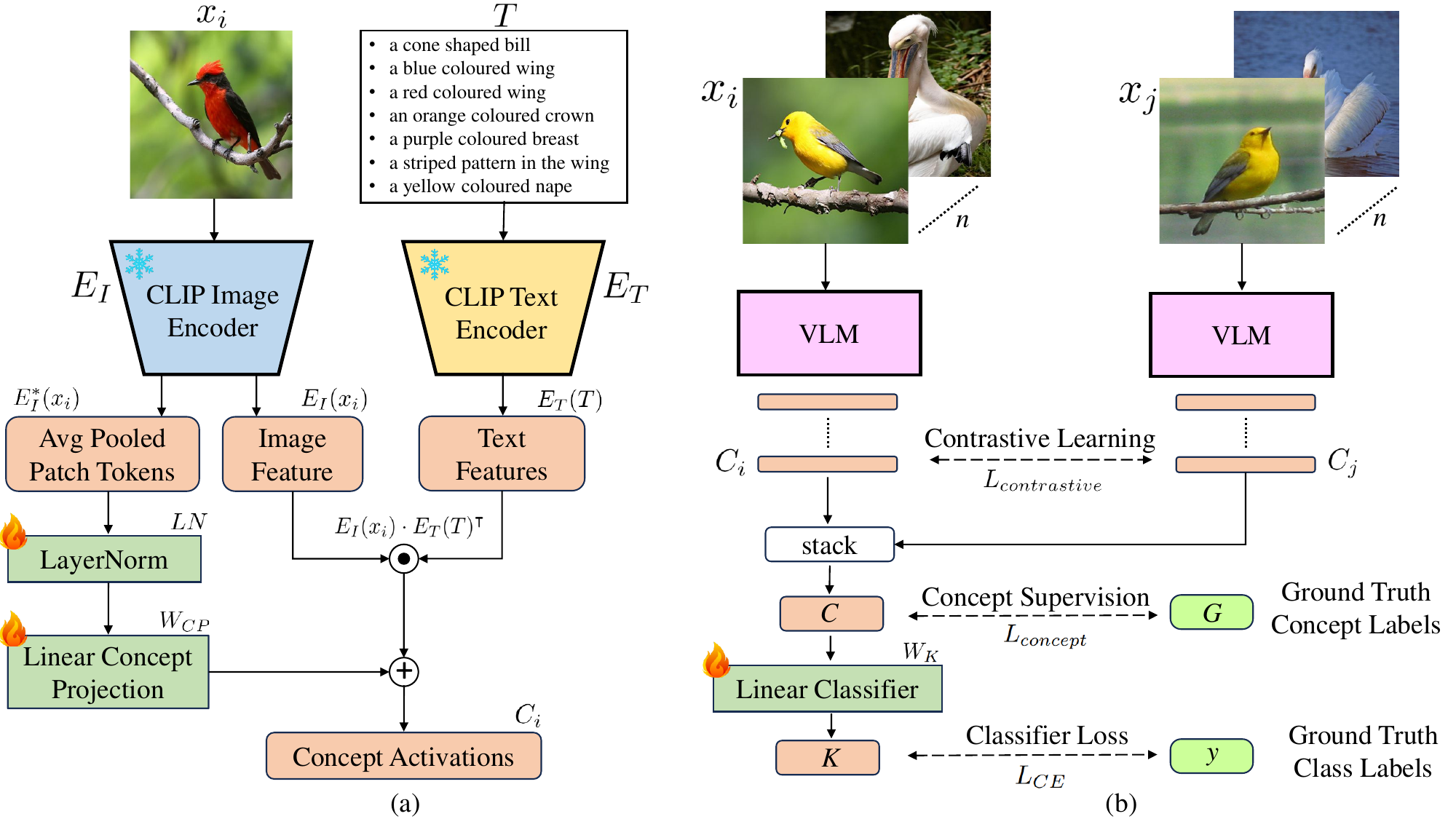}
  \caption{(a) Modified CLIP (VLM) with concept projection layer which enhances the raw concept scores. (b) Architecture of VL-CBM and an overview of the Contrastive Semi-Supervised (CSS) learning method.}
  \label{fig:architecture}
  \vspace{-1.5em}
\end{figure*}

\subsection{Enhancing Faithfulness of VL-CBMs}
To improve concept alignment in VL-CBM we propose a novel Contrastive Semi-Supervised (CSS) learning method that leverages a few labelled concept examples per class. Figure~\ref{fig:architecture} illustrates the overall framework. As shown in Table~\ref{tab: poor concept alignment}, the frozen CLIP model achieves a low yet non-trivial concept accuracy. We enhance this by adding a learnable linear concept projection layer (see Figure~\ref{fig:architecture}a) which improves the CLIP concept scores by directly predicting $c$ concepts from the average pooled visual patch tokens as,
\begin{equation}
    \label{eqn:1}
    C_i = E_I(x_i) \cdot E_T(T)^{\intercal} + LN(E_I^*(x_i)) \cdot W_{CP},
\end{equation}

\vspace{-0.2em}
\noindent where $E_I(x_i) \in \mathbb{R}^{1 \times 512}$ is the image feature, $E_T(T) \in \mathbb{R}^{c \times 512}$ is the text feature for $c$ concepts, $E_I^*(x_i) \in \mathbb{R}^{1 \times 768}$ denotes the average pooled patch tokens from the visual encoder, $ W_{CP} \in \mathbb{R}^{768 \times c}$ is the learnable concept projection layer and $LN$ denotes layer normalization. From the concept scores $C_i  \in \mathbb{R}^{1 \times c}$, a linear classifier $W_K \in \mathbb{R}^{c \times k}$ makes class predictions $K_i = C_i \cdot W_K$ for the image $x_i$, where $K_i \in \mathbb{R}^{1 \times k}$ denotes the class logits for the $k$ classes.

Now, the model must be optimized to enhance both concept and classification accuracies. To improve concept accuracy, the concept projection layer can be trained with a few concept labels (semi-supervision), which refines the weak concept scores of CLIP and aligns the predicted concepts with the ground truth. However, one needs to effectively leverage this sparse (concept) supervisory signals and extend them to other unlabelled examples. Contemporary unimodal works~\cite{hu2024semi} compare the features of labelled and unlabelled examples in the cosine space, and then assign pseudo concept labels for unlabelled examples using KNN algorithm.  
But, this necessitates additional computation steps (KNN) for each mini-batch. We instead propose to bootstrap the sparsely supervised concept scores to other unlabelled samples using contrastive learning. This method promotes consistent concept scores for samples sharing `similar' concepts (positive pairs) and discriminates concept scores for samples with `disjoint' concepts (negative pairs). 
For all datasets in our paper, the similar concept pairs come from intra-class samples and the disjoint concept pairs come from inter-class samples. Furthermore, discriminating the scores of inter-class samples make classes more separable in the concept space, thus aiding in classification.

To achieve this (see Figure~\ref{fig:architecture}b), we perform pairwise sampling such that each training sample consists of a pair of images $(x_i,x_j)$ from the same class, while every other sample in the mini-batch belongs to different classes. There are $n$ samples ($2n$ images) in each mini-batch, and only a subset of these images will be concept supervised. For these image pairs, we compute the concept scores $(C_i,C_j)$ and class predictions $K = [K_i,K_j]~\footnote{The square brackets denote the tensors are stacked across batch dimension and no longer need to be in pairs.}$. Let $g = [g_i,g_j]$ and $y = [y_i,y_j]$ be the corresponding ground-truth concept labels and class labels. We jointly train the concept projection layer and the linear classifier to minimize the following loss objective.

\begin{flalign}
\label{eqn:trinity-loss}
    &L = L_{contrastive} + L_{CE} + L_{concept}, \\ \nonumber
    &L_{contrastive} = \frac{1}{n} \sum_{i,j \in n} -\log\frac{e^{sim(C_i,C_j)/\tau}}{\sum_{m \neq i} e^{sim(C_i,C_m)/\tau}}, \\ \nonumber
    &L_{CE} = \frac{1}{2n} \sum_{l \in 2n} L_{ce}(K_l,y_l), \\ \nonumber
    &L_{concept} = \frac{1}{2n} \sum_{l \in 2n} L1(\gamma \cdot (s(C_l) - s(G_l))),\ \ \textrm{if}\ \ |G_l| \neq 0, \\ \nonumber
    &L_{concept} = 0,\ \ \textrm{if}\ \ |G_l| =  0,
\end{flalign}

\noindent where $sim$ is the cosine similarity metric, $\tau$ is the temperature parameter, $L_{ce}$ and $L1$ is the standard cross entropy loss and mean absolute error, $s$ is the softmax normalization and $\gamma$ is a constant that increases the magnitude of the $L1$ loss function. $G_l$ represents the ground truth concept label for the $l$-th entry. $l$ denotes the index of the concept score, class logit, and label tensor within the mini-batch, after stacking along the batch dimension.

\begin{itemize}
    \item The \textbf{contrastive loss} ($L_{contrastive}$) optimizes $W_{CP}$ and encourages the concept scores of the same classes (positive pairs) to be consistent and discriminates it from the other classes (negative pairs).
    \item The \textbf{cross entropy loss} ($L_{CE}$) trains the linear classifier $W_K$ to make class predictions from the concept scores.
    \item The \textbf{concept loss} ($L_{concept}$) optimizes $W_{CP}$ to align the concept predictions with the ground truth concept labels and it primarily improves the concept accuracy. The concept loss is enabled if the concept labels are available ($|G_l| \neq 0$) or else it becomes zero.
\end{itemize}


\begin{table*}[t]
  \centering
  \vspace{-0.5em}
  \begin{adjustbox}{width=0.75\textwidth}
  \begin{tabular}{c|cc|cc|cc|c}
    \toprule
    \multirow{2}{*}{\textbf{Model}} & \multicolumn{2}{c}{\textbf{CUB}} & \multicolumn{2}{c}{\textbf{RIVAL}} & \multicolumn{2}{c}{\textbf{AwA2}} & \textbf{WBCAtt} \\
     & ~Class  & ~Concept  & ~Class  & ~Concept  & ~Class  & ~Concept  & Attribute  \\
    \midrule
    ViT-B/16 (Black Box) & ~\underline{79.21} & NA & ~\textbf{99.47} & NA & ~~\textbf{94} & NA & ~\textbf{66.11} \\
    CLIP VL-CBM & ~75.84 & 24.43 & ~95.63 & 58.85 & ~90.14 & 49.02 & 42.56 \\
    \textbf{CSS VL-CBM (Ours)} & ~\textbf{81.45} & \textbf{63.53} & ~\underline{98.47} & \textbf{77.48} & ~~\underline{93.2} & ~\textbf{81.13} & ~\underline{64.19} \\
    \bottomrule
  \end{tabular}
  \end{adjustbox}
  \caption{Improving Concept Alignment in VL-CBM using Contrastive Semi-Supervised (CSS) learning. We report the classification, concept, and WBC attribute accuracies (in \%) as Class, Concept, and Attribute. The best performance is \textbf{bolded}, and the second-best is \underline{underlined}. Note that WBCAtt uses the PLIP model.}
  \label{tab: improving concept alignment}
  \vspace{-0.5em}
\end{table*}

\section{Experiments}

\subsection{Experimental Setting}

\noindent \textbf{Dataset.} We validate the effectiveness of our Contrastive Semi-Supervised (CSS) method in improving concept alignment on the CUB~\cite{wah2011caltech}, RIVAL~\cite{moayeri2022comprehensive}, AwA2~\cite{xian2018zero} datasets, and in a medical setting on the White Blood Cell Attribute (WBCAtt) dataset~\cite{tsutsui2023wbcatt}. All the above datasets provide an expert-defined concept set and the corresponding ground-truth concept labels, which is crucial for our evaluation.

\vspace{0.2em}
\noindent \textbf{Model.} We use the PLIP ViT-B-16~\cite{huang2023visual} model available in Hugging Face for experiments on WBCAtt dataset and use the OpenCLIP ViT-B-16 model~\cite{ilharco_gabriel_2021_5143773} for experiments on all other datasets, unless otherwise specified.

\vspace{0.2em}
\noindent \textbf{Training and Evaluation.} Following the literature~\cite{koh2020concept,fong2017interpretable}, we use concept accuracy to quantify concept alignment. We use Adam optimizer to train all our models and evaluate them with classification accuracy and concept accuracy metrics. We also conduct a distributional analysis of the VL-CBM concept scores to show the concept alignment explicitly.


\subsection{Improving Concept Alignment}

We empirically demonstrate the effectiveness of our Contrastive Semi-Supervised (CSS) approach in improving concept alignment for the CUB, RIVAL, AwA2, and WBCAtt datasets in Table \ref{tab: improving concept alignment}.

\vspace{0.2em}
\noindent \textbf{Bird and Animal classification.} We validate our CSS method with concept accuracy and classification accuracy metrics for the CUB, RIVAL, and AwA2 datasets. We first train a black-box image classifier with frozen CLIP vision encoder (ViT-B/16) as the backbone to establish baseline classification performance. The black-box classifier directly predicts the classes from image features without any concept bottlenecks. We then train a CLIP VL-CBM~\cite{oikarinenlabel,wang2023learning} classifier with expert-defined text concepts. From the table, we can see that the classification performance of CLIP VL-CBM is on par with the black-box classifier, however, it has a lower concept accuracy.

Hence, to improve the faithfulness of VL-CBM, we add the concept projection layer and optimize the model with our CSS objective (CSS VL-CBM). Our CSS method not only substantially increases the concept accuracy, it also closes the gap with black-box methods on image classification performance. Especially for the CUB dataset, our CSS method surpasses the black-box classifier. The \textit{``number of concept labels per class''} used to train the CSS VL-CBM model for the above datasets are as CUB - 9 labels (30\%), RIVAL - 8 labels (0.1\%), and AwA2 - 10 labels (1.4\%). The values within brackets denote the \textit{``concept labels used for training as a percentage of the total concept labels provided by the dataset''}. 
The results show that even with such small percentages of concept labels, our CSS method can substantially increase the concept accuracy.
The enhanced and faithful concept scores after improving concept alignment with CSS is presented in Figure~\ref{fig:concept visualizations} (CSS Concept Scores).

\vspace{0.2em}
\noindent \textbf{WBC Attribute prediction.} WBCAtt contains White Blood Cell (WBC) images annotated with 31 ``cell morphology'' attributes. 
These attributes aid in diagnosis of a variety of conditions in Hematology and Pathology. So, we directly evaluate the models on their attribute prediction accuracy and do not perform any image classification. While contemporary works~\cite{yuksekgonul2023posthoc,yang2023language} use the HAM10000~\cite{tschandl2018ham10000} dataset to study VL-CBM in the medical domain, the availability of attribute labels in WBCAtt makes the latter suitable for our work. For our evaluation, we train the PLIP model (a VLM for Pathology AI), with 50\% of the attribute labels.  From the table, we can see that our CSS method substantially increases the attribute prediction accuracy and closes the gap with black-box attribute predictor.


\begin{figure}[t]
  \centering
  \includegraphics[width=0.95\linewidth]{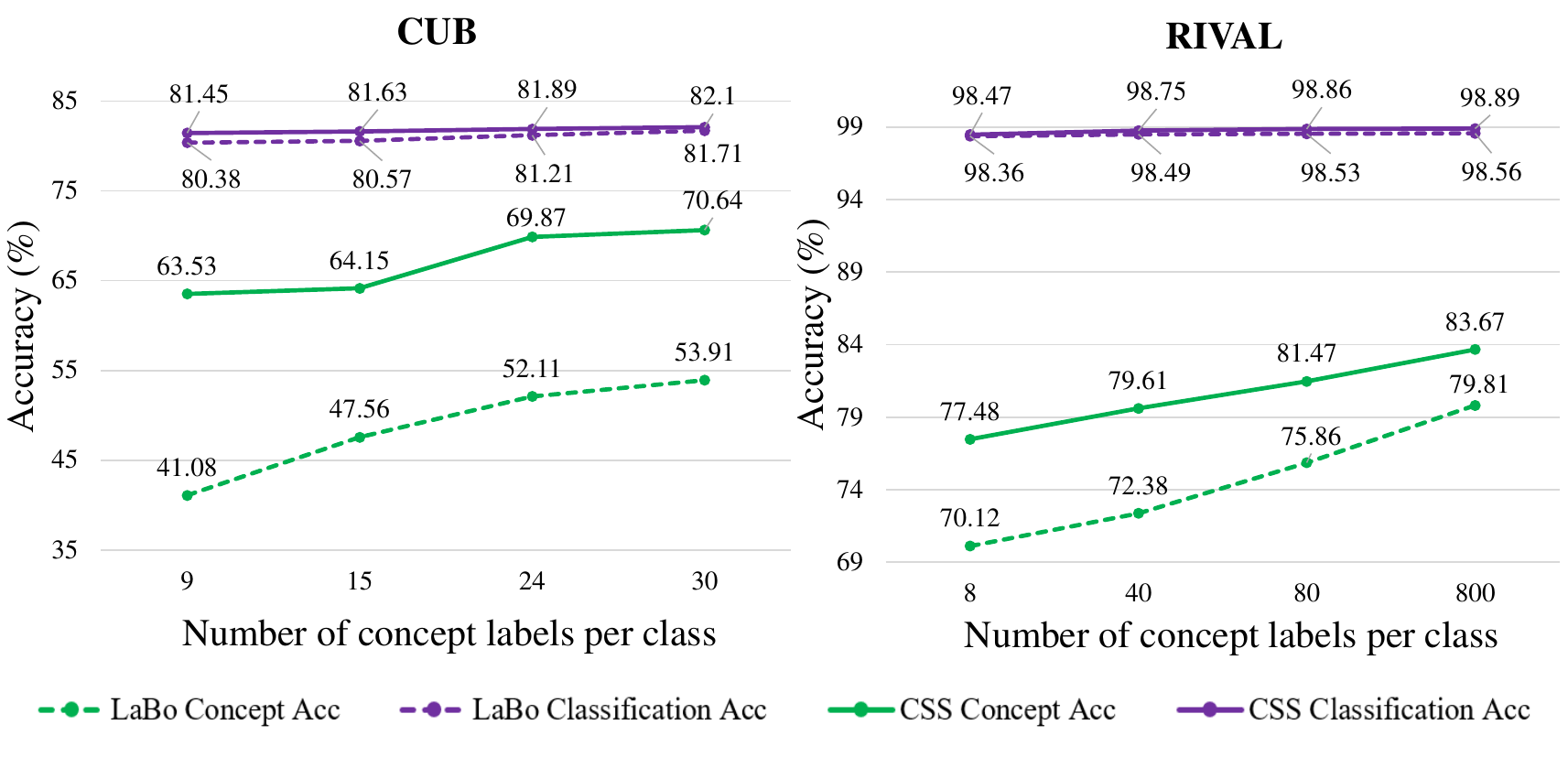}
  \vspace{-0.5em}
  \caption{Impact of the number of concept labels on CSS VL-CBM and LaBo.}
  \label{fig:num cl vs acc}
  \vspace{-0.5em}
\end{figure}

\vspace{0.3em}
\noindent \textbf{Impact of the number of concept labels.} Since our approach is semi-supervised, we further investigate the impact of the number of human-annotated concept labels on CSS VL-CBM performance, 
as shown in Figure~\ref{fig:num cl vs acc}. In this study, we also compare our method with the closest work, LaBo, where we fine-tune the final layernorm and visual projection layers of the CLIP model (which closely resembles training our concept projection layers). Both models are trained with expert concepts, and we progressively increase the number of concept labels used for training on the CUB and RIVAL datasets. From the figure, we observe that both approaches achieve similar classification accuracies. However, our CSS VL-CBM model attains relatively higher concept accuracies, especially when fewer concept labels are available. We attribute this to the use of the concept projection layer (which enhances raw CLIP scores) and the application of contrastive learning.


\begin{table}[t]
    \centering
    \begin{adjustbox}{width=0.48\textwidth}
    \begin{tabular}{cccc}
        \toprule
        \textbf{CBM Method} & \textbf{Concepts} & \textbf{Source} & \textbf{Class Acc (\%)} \\
        \midrule
        LaBo~\cite{yang2023language} & 10,000 & LLM & 81.9 \\
        SparseCBM~\cite{semenov2024sparse} & 926 & LLM & 80.02 \\
        CDL~\cite{zang2024pre} & 400 & LLM & \underline{83.4} \\
        LM4CV~\cite{yan2023learning} & 400 & LLM & 81.4 \\
        \textbf{CSS VL-CBM (Ours)} & 321 & Expert & \textbf{83.89} \\
        \bottomrule
    \end{tabular}
    \end{adjustbox}
    \vspace{-0.5em}
    \caption{Comparison of CSS VL-CBM with contemporary methods on CUB dataset.}
    \label{tab: cub comparison}
    \vspace{-0.5em}
\end{table}

\subsection{Comparison with other methods}

\noindent \textbf{CUB.} We compare our Contrastive Semi-Supervised VL-CBM method with contemporary VLM-based interpretable models on the CUB dataset, which is extensively used by the Explainable AI community to study CBMs. Following the literature, we adopt the CLIP ViT-L-14 model for this comparison. Note that this increases the number of trainable parameters in the concept projection layer. The results, presented in Table \ref{tab: cub comparison}, show that our CSS model achieves strong classification performance despite having a smaller number of bottleneck concepts. Since other methods use LLM-generated concepts without corresponding concept labels, concept accuracy is not reported for this comparison.


\begin{table}[t]
  \centering
  \begin{adjustbox}{width=0.38\textwidth}
  \begin{tabular}{ccc}
    \toprule
    \textbf{CBM Method} & \textbf{Class Acc (\%)} & \textbf{Concept Acc (\%)} \\
    \midrule
    Text2Concept & 95.38 & \_ \\
    CSS (Ours) & \textbf{98.86} & 72.84 \\
    \midrule
    PCBM & 88 & 71.9 \\
    ECBM & \underline{91.2} & \textbf{85.84} \\
    CSS (Ours) & \textbf{93.2} & \underline{81.3} \\
    \bottomrule
  \end{tabular}
  \end{adjustbox}
  \caption{Comparison of CSS VL-CBM with other methods on RIVAL (top) and AwA2 (bottom) datasets.}
  \label{tab: rival and awa2 comparison}
  \vspace{-0.5em}
\end{table}

\vspace{0.3em}
\noindent \textbf{RIVAL} and \textbf{AwA2.} We further compare CSS VL-CBM with contemporary methods on the animal classification datasets in Table \ref{tab: rival and awa2 comparison}. For the RIVAL dataset, CSS achieves higher classification performance compared to Text2Concept~\cite{moayeri2023text}. For the AwA2 dataset, CSS achieves higher classification accuracy compared to other methods but slightly lower concept accuracy compared to ECBM. It is important to note that PCBM~\cite{kim2023probabilistic} and ECBM~\cite{xu2023energy} do not use ``text" concepts and optimize the vision backbone during training.


\begin{table}[t]
  \centering
  \begin{adjustbox}{width=0.4\textwidth}
  \begin{tabular}{ccc}
    \toprule
    \textbf{Metric} & \textbf{CSS Scores} & \textbf{CLIP Scores} \\
    \midrule
    Truthfulness$^*$ ($\downarrow$) & \textbf{3.22} & 14.04 \\
    Sparseness$^\#$ ($\downarrow$) & \textbf{0.08} & 0.19 \\
    Discriminability$^\#$ ($\uparrow$) & \textbf{7.04} & 3.28 \\
    \bottomrule
  \end{tabular}
  \end{adjustbox}
  \caption{Distributional analysis of CSS vs CLIP concept scores for CUB test. The results are averaged over instances$^*$ and classes$^\#$.}
  \label{tab: distributional analysis}
  \vspace{-0.5em}
\end{table}

\subsection{Distributional Analysis of Concept Space}
\begin{figure}[t]
  \centering
  \includegraphics[width=0.7\linewidth]{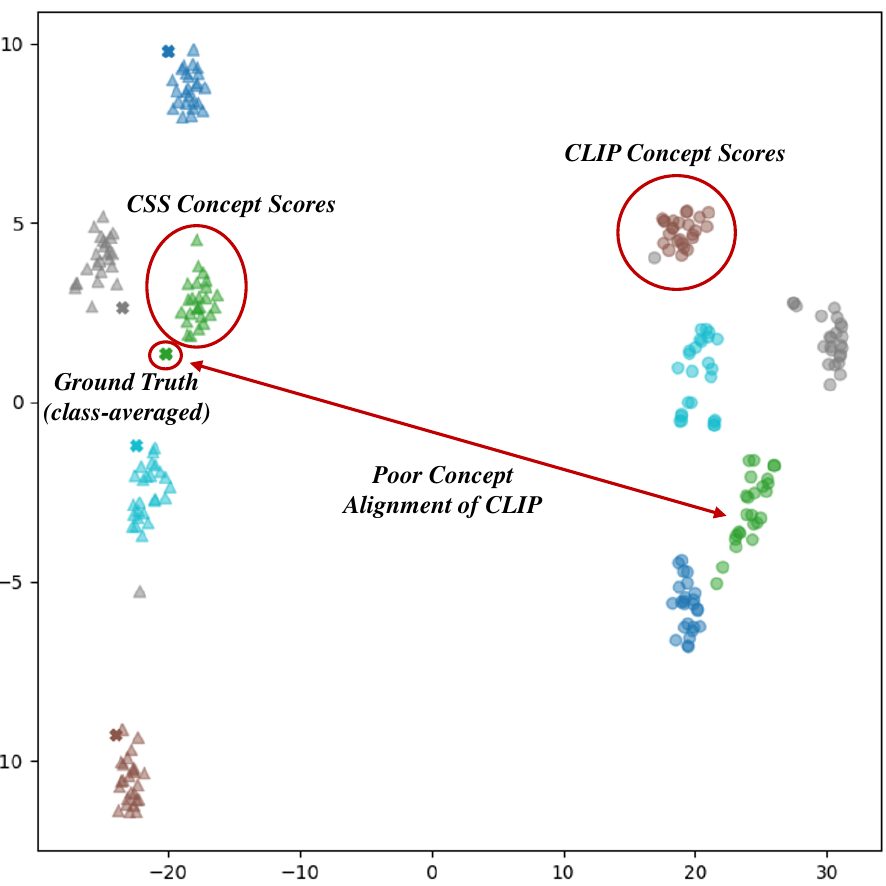}
  \vspace{-0.5em}
  \caption{t-SNE visualization of CSS (triangles) vs CLIP (circles) concept scores of 5 random classes for CUB test.}
  \label{fig:distributional Analysis}
  \vspace{-0.5em}
\end{figure}

In addition to concept accuracy metric, we conduct a distributional analysis of the VL-CBM concept scores to explicitly demonstrate the effectiveness of our CSS method in improving concept alignment. For quantitative assessment, we extract the concept scores ($\mathbb{R} ^{5790 \times 312}$) from both CLIP and our CSS model for the CUB test set. We report three metrics: \textbf{\textit{truthfulness}:} defined as the L2-distance between predicted and ground-truth concepts; \textbf{\textit{sparseness}:} measures the intra-class standard deviation of the predicted concepts; and \textbf{\textit{discriminability}:} defined as the L2-distance between inter-class concept score clusters. 
The results in Table~\ref{tab: distributional analysis} show that the proposed CSS model has improved concept alignment compared with the CLIP model. To further visually compare the difference, Figure~\ref{fig:distributional Analysis} depicts a t-SNE visualization of CSS (triangles) vs. CLIP (circles) concept scores of 5 random classes on the CUB test set. Our CSS method makes the concept clusters more dense and more separable in the concept space, enhancing concept alignment by bringing them closer to the ground truth.

\section{Intervention}

When a CBM makes an error, one can inspect the incorrect predictions and manually adjust the classifier weights or modify the concepts to rectify these errors, thereby improving the overall accuracy. This process is known as ``intervention" or ``model editing." Traditional intervention approaches are typically performed at the instance level~\cite{oikarinenlabel,yuksekgonul2023posthoc,koh2020concept}, where a few erroneous images are arbitrarily chosen for intervention. However, for fine-grained multi-class classification problems like CUB, which have many classes (some bird species are visually similar) and relatively few samples per class, selection of error images become non-trivial. Therefore, we propose a class-level intervention procedure where we first identify the confounding classes and then choose the error images from these classes for intervention. Confounding classes are bird species that are visually similar yet semantically different, and the model makes the most errors for these challenging classes. This class-level intervention is carried out in the following steps.

\vspace{0.2em}
\noindent\textbf{Step 1: Identify the confounding classes.} We compute a $K \times K$ error matrix by evaluating the trained CSS VL-CBM model on the test set, where  $K$ is the number of classes. The error matrix provides the error distribution for each class, and the matrix element with the highest value indicates the confounding class pairs. In our experiments, we identified two pairs of confounding classes: \textit{California Gull - Western Gull} and \textit{Common Tern - Arctic Tern}. The total number of classification errors for these confounding classes before intervention is presented in Table \ref{tab: intervention}.

\vspace{0.2em}
\noindent\textbf{Step 2: Inspecting the concept space.} Next, we inspect the concept scores of the error images in these confounding class pairs. We find that the concept score distribution of the confounding classes is nearly identical, making it difficult to discriminate between them. Even the averaged ground-truth concept labels for the confounding classes are highly similar. Our model suffers from Type-2 error~\cite{oikarinenlabel}, meaning the number of bottleneck concepts and their score distributions are insufficient to effectively discriminate the samples.


\begin{table}[t]
  \centering
  \vspace{-0.5em}
  \begin{adjustbox}{width=0.48\textwidth}
  \begin{tabular}{ccc}
    \toprule
     & \textbf{Before Interv.} & \textbf{~~After Interv.} \\
    \midrule
    Total error for CG (30) & 27 & 14 \\
    Total error for WG (30) & 15 & 9 \\
    CG misclassified as   WG & 13 & 4 \\
    WG misclassified as   CG & 3 & 1 \\
    \midrule
    Total error for CT (30) & 21 & 11 \\
    Total error for AT (29) & 3 & 2 \\
    CT misclassified as   AT & 12 & 6 \\
    AT misclassified as   CT & 0 & 0 \\
    \midrule
    Classification Accuracy (\%) & \textbf{81.45} & \textbf{82.57} \\
    \bottomrule
  \end{tabular}
  \end{adjustbox}
  \caption{Class-level intervention of CSS VL-CBM. The confounding classes are as California Gull (CG), Western Gull (WG), Common Tern (CT), and Artic Tern (AT). The value within brackets denote the total number of test samples for that class.}
  \label{tab: intervention}
  \vspace{-1.5em}
\end{table}


\vspace{0.2em}
\noindent\textbf{Step 3: Expanding the concept set.} To mitigate the Type-2 error and to improve the overall performance, we expand the concept set by leveraging the concepts provided by LaBo~\cite{yang2023language}. Note that the new concepts to be added should be visually groundable by the CLIP model and hence we use the ``filtered'' concept set of LaBo. We choose $top-32$ concepts per confounding class and expand the concept set from the original 312 concepts to 440 concepts. Formally, let $T'$ be the new set of 128 text concepts and $C'_i = E_I(x_i) \cdot E_T(T')^{\intercal}$ be the CLIP concept scores for it. Now the concept scores for all the 440 concepts for an image $x_i$ is given by $C'=$ Concat($C_i,C'_i$). During this intervention procedure we do not train a new concept projection layer for the expanded concept set.

\vspace{0.2em}
\noindent\textbf{Step 4: Training the classifier.} We add a new linear classifier $W'_K \in \mathbb{R}^{128 \times 4}$ that directly predicts the confounding classes as $K' = C' \cdot W'_K$. Then we append zeros for the non-confounding classes and add them to the original class predictions as $K'_i = K_i + K'$. We finally train the linear classifiers $W_K$ and $W'_K$ simultaneously on the CUB train set in a standard supervised setting with cross-entropy loss $L_{ce}$. The total number of classification errors after this class-level intervention procedure is presented in Table \ref{tab: intervention}. We observe a substantial decrease in errors for the confounding classes with an overall increase in the classification performance.

\section{Limitations}

Despite the numerous advantages, VL-CBMs have a few potential limitations.

\vspace{0.2em}
\noindent \textbf{Ineffable concepts.} While natural language offers a convenient way to build interpretable models with high-level abstract concepts, its expressive power can be limited. The subtle visual cues required for certain classification tasks, such as face recognition, might be difficult to articulate in words.

\vspace{0.2em}
\noindent \textbf{Unknown concepts.} VL-CBMs assume that the complete set of salient concepts required for solving the classification task is known before training. However, this assumption is not always feasible. In such cases, models that can ``discover" unknown concepts~\cite{shang2024incremental} during the course of training are more suitable.

\vspace{0.2em}
\noindent \textbf{Locality Faithfulness.} Recent works have observed that even when CBMs predict the concept correctly, they often fail to \textit{`respect localities'}~\cite{raman2024concept}. This suggests that the concept predictor might rely on spurious correlations instead of relevant visual/spatial features. Such a study on locality faithfulness in VL-CBMs is largely unexplored.


\section{Conclusion}
In this paper, we investigated the faithfulness of VL-CBM for expert concepts and found that VLMs like CLIP have poor concept alignment. Hence, we proposed a novel Contrastive Semi-Supervised approach that improved concept alignment with fewer human-annotated concept labels. Our work will primarily aid the Explainable AI community in building interpretable models that align with human preferences (concepts) yet require minimal supervision. We also introduced a class-level intervention procedure for fine-grain classification problems that reduced the error for the confounding classes to increase the overall performance of the model.


\bibliography{aaai25}

\end{document}